\documentclass[10pt,twocolumn,letterpaper]{article}

\usepackage{cvpr}              % To produce the CAMERA-READY version

\usepackage{multirow}

\usepackage{booktabs}
\usepackage{amsmath,amssymb,amsthm}
\newtheorem{proposition}{Proposition}

\usepackage{algorithm}
\usepackage{algorithmic}
\usepackage[table]{xcolor}
\usepackage{colortbl}
\definecolor{papergreen}{cmyk}{0.91,0,0.88,0.12}

\definecolor{cvprblue}{rgb}{0.21,0.49,0.74}
\usepackage[pagebackref,breaklinks,colorlinks,allcolors=cvprblue]{hyperref}

\def\confName{CVPR}
\def\confYear{2027}

\title{EM-SDTrack: Efficient Multi-Scale Spike-driven Tracking for Event Cameras}

\author{Yimeng Shan$^{1,2}$, Haodi Wu$^{1}$, Wei Zheng$^{1}$, 
Xiaowei Ye$^{3}$, Shuai Wang$^{1,2}$,\\ Yu Liang$^{1,2}$, Wei Hua$^{3}$, Chenxiao Dou$^{3}$, Malu Zhang$^{\dag1,2}$\\
$^1$University of Electronic Science and Technology of China, China \\
$^2$Shenzhen Loop Area Institute, China\\
$^3$China Nanhu Academy of Electronics and Information Technology, China\\
\texttt{yimengshan2001@gmail.com} \\
Code: \url{https://github.com/YmShan/EM-SDTrack}
}

\begin{document}
\maketitle

\begin{abstract}
Event-driven spiking neural networks (SNNs) match the sparse spatiotemporal
output of event cameras, making spike-driven trackers highly attractive for
neuromorphic edge deployment. However, events reveal a target only through
their spatial distribution, whose extent varies more than tenfold within one
sequence. Event tracking therefore demands multi-scale representation, which
existing spike-driven trackers lack. Transferring multi-scale schemes from
artificial neural networks either breaks spike-driven inference with
floating-point multiplication or repeats backbone computation across input
scales. To address this, we propose Spike-driven Multi-Scale Interaction
(SDMI), which performs cross-scale interaction in a single spiking forward
pass. Its core, Dyadic Spike Sampling (DSS), restricts sampling coordinates
to dyadic fractions and thus requires only shifts and additions. The
resulting M-SDTrack is fully spike-driven at inference. For tighter edge
budgets, we further propose Polarity-Privileged Distillation (PPD), which
distils M-SDTrack into EM-SDTrack. PPD treats the event counts cancelled by
the voxel representation as privileged
information. Its core constraint, Polarity-Guided Curvature Distillation
(PGCD), aligns the curvature of the template--search relation under
polarity-guided perturbations. On FE108, FELT and VisEvent, M-SDTrack
achieves state-of-the-art accuracy with an AUC of 65.1, 41.7 and 42.0.
EM-SDTrack retains 99\% of the accuracy of its teacher M-SDTrack-Tiny with
27\% of its energy and 18\% of its parameters. On the Lynxi KA200 neuromorphic chip, it
runs losslessly in real time at 116~FPS and 0.12~W.
\end{abstract}

\section{Introduction}

\begin{figure}[t]
  \centering
  \includegraphics[width=0.45\textwidth]{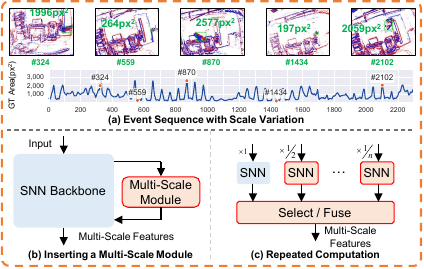}
  \caption{Scale variation in event tracking and the cost of existing
remedies. (a) Ground-truth box area of one target along a sequence, which
ranges from 197 to 2577 pixels. (b) A multi-scale module inserted into an
SNN backbone adds floating-point sampling and weighting, so inference is no
longer spike-driven. (c) Repeating the network at several input scales
stays spike-driven but multiplies the computation. SDMI avoids both costs.}
  \label{visual}
\end{figure}

Event cameras emit asynchronous events only at pixels where brightness
changes, so their output is sparse and rich in spatiotemporal information~\cite{gallego2020event}.
Spiking neural networks (SNNs) match this sparsity by computing with binary
spikes and replacing multiply-accumulate operations with accumulations~\cite{maass1997networks,shanspiking}.
SDTrack exploits this match and builds a tracking pipeline entirely from
SNNs on a one-stream architecture~\cite{shan2026sdtrack}. Such spike-driven
trackers are intended for neuromorphic chips, so their practical value
depends on deployment on edge devices~\cite{davies2018loihi,merolla2014million}. Deployment imposes two constraints.
The inference path must remain spike-driven and admit no floating-point
multiplication. The model must also fit the storage and weight-bandwidth
limits of edge devices.

The first obstacle under these constraints is scale variation, which event
data make harder than in RGB tracking. Events carry no texture or colour, so
a target appears only as the spatial distribution of events along its moving
edges. This distribution is the sole cue for localisation, yet its extent
changes drastically with the relative motion between camera and target.
Within one sequence, the visible area of a target varies from a few hundred
to several thousand pixels (Fig.~\ref{visual}a). A tracker must therefore
represent the target at multiple scales. Existing spike-driven trackers lack
this ability, because their hierarchical
backbones~\cite{yao2024spikev2,yao2025scaling} pass only the last
single-resolution feature to the head.

% Multi-scale perception is mature in artificial neural networks, so the
% question is the deployment cost of transferring it~\cite{liu2020deep}. Two routes prevail, and
% each hinders edge deployment in its own way (Fig.~\ref{visual}b, c). The
% first inserts modules such as feature pyramids and multi-scale deformable
% attention into the network. Their sampling, interpolation and weighting rely
% on floating-point multiplication, so inference is no longer spike-driven~\cite{lin2017feature,zhu2020deformable,xia2024vit}. The
% second runs the backbone repeatedly at several input scales and fuses the
% branches, which multiplies the computation with the number of branches. To
% our knowledge, no method performs cross-scale interaction within a single
% spiking forward pass~\cite{chen2016attention}.

Multi-scale perception is mature in artificial neural networks, so the
question is the deployment cost of transferring it~\cite{liu2020deep}. Two routes prevail, and
each hinders edge deployment in its own way (Fig.~\ref{visual}b, c). The
first inserts modules such as feature pyramids and multi-scale deformable
attention into the network. Their sampling, interpolation and weighting rely
on floating-point multiplication, so inference is no longer spike-driven~\cite{lin2017feature,zhu2020deformable,xia2024vit}. The
second keeps inference spike-driven but runs the network repeatedly at
several input scales and fuses the branches, which multiplies the
computation with the number of branches. To
our knowledge, no method performs cross-scale interaction within a single
spiking forward pass~\cite{chen2016attention}.

We propose Spike-driven Multi-Scale Interaction (SDMI) to fill this gap.
SDMI builds a multi-scale buffer from the three convolutional stages. Before
each Transformer stage, the search tokens read cross-scale context from this
buffer. The core of SDMI, Dyadic Spike Sampling (DSS), restricts the
continuous sampling coordinates of bilinear interpolation to $b$-bit dyadic
fractions (Algorithm~\ref{alg:dss}). Every interpolation weight then becomes
an integer over a power of two, so each weight--value product reduces to
shifts and additions (Proposition~\ref{prop:dss}). The aggregation weights
are likewise spikes, so the interaction involves no floating-point
multiplication. The resulting M-SDTrack is therefore fully spike-driven at
inference.

For tighter edge budgets, we further propose Polarity-Privileged
Distillation (PPD), which distils M-SDTrack into EM-SDTrack. The student
keeps the topology of the teacher at about half the channel width, retains two
Transformer blocks after progressive shrinking~\cite{cui2023mixformerv2},
shrinks the head accordingly and has no SDMI module. Its input is the voxel representation, which underlies
the best-performing spike-driven trackers (Appendix~A.3) but
cancels positive and negative events at the same pixel~\cite{yang2026fully}.
These collisions fall exactly where a tracker needs evidence. PPD therefore
treats the cancelled counts as privileged information that decides where and
in which direction teacher and student are compared. Its core constraint,
Polarity-Guided Curvature Distillation (PGCD), matches the curvature of the
template--search relation under perturbations guided by these counts.
Counts, adapter and perturbations exist only in training, so EM-SDTrack
keeps the plain voxel input and its latency at inference. Our
contributions are summarised as follows.
\begin{itemize}
\item We propose SDMI, which performs cross-scale interaction in a single
spiking forward pass. Its core, DSS, restricts sampling coordinates to
dyadic fractions, so sampling costs only shifts and additions. The resulting
M-SDTrack is fully spike-driven at inference.
\item We propose PPD, which distils M-SDTrack into EM-SDTrack. It treats
the event counts cancelled by the voxel representation as privileged
information. Its core constraint, PGCD, matches the curvature of the
template--search relation under polarity-guided perturbations.
% \item On three benchmarks, M-SDTrack achieves state-of-the-art accuracy.
% EM-SDTrack retains 99\% of the accuracy of its teacher M-SDTrack-Tiny with
% 18\% of its parameters and 27\% of its energy. On the Lynxi KA200 neuromorphic chip, it
% runs losslessly in real time at 116~FPS and 0.12~W.
\item On three benchmarks, M-SDTrack achieves state-of-the-art accuracy.
EM-SDTrack retains 99\% of the accuracy of M-SDTrack-Tiny with
18\% of its parameters and 27\% of its energy. On the Lynxi KA200 neuromorphic chip, it
runs losslessly in real time at 116~FPS and 0.12~W.
\end{itemize}

\section{Related work}
\subsection{Event-based tracking}
Early event trackers rely on traditional artificial neural networks
(ANNs) and exploit the spatiotemporal properties of
events~\cite{messikommer2025data}, their temporal
resolution~\cite{wang2026towards,yang2019dashnet} or RGB--event
fusion~\cite{zhu2023cross,wang2024long}. These trackers, however,
discard event sparsity. Zhang et al. first exploit it with ANN--SNN
hybrids~\cite{zhang2022spiking,zhang2025spiking}. SDTrack then builds
the first pipeline entirely from SNNs~\cite{shan2026sdtrack}. AdaS
improves its training~\cite{zhouadas}, SpikeFET regularises
cross-timestep feature similarity~\cite{yang2026fully} and TP-Spikformer
sparsifies its computation~\cite{wei2026tp}. None of
them, however, shrinks the tracker for edge storage and bandwidth
budgets. The PPD we propose fills this gap.

\subsection{Multi-scale feature interaction}
Multi-scale perception, absent from spike-driven trackers, is well
studied in ANNs. Chen et al. fuse one network's outputs at several
scales~\cite{chen2016attention}. FPN builds a
top-down pyramid from one backbone pass~\cite{lin2017feature}. Deformable
DETR bilinearly samples multi-scale features at predicted
offsets~\cite{zhu2020deformable}. ViT-CoMer couples a
multi-receptive-field pyramid with a ViT by deformable
attention~\cite{xia2024vit}. Hierarchical SNN
backbones~\cite{yao2024spikev2,yao2025scaling,zhou2024qkformer,fang2026spiking}
produce multi-resolution features without cross-scale fusion. Shan et al.
add such fusion to SNNs, weighting multi-receptive-field branches by
temporal and channel attention~\cite{shan2025advancing}. Each scheme,
however, hinders edge deployment. Multi-scale input repeats the backbone
computation, and the fusion modules, spiking ones included, weight
features by real-valued multiplication. The SDMI we propose performs
cross-scale interaction in one backbone pass without floating-point
multiplication.

\begin{figure*}[ht]
  \centering
  \includegraphics[width=1.0\textwidth]{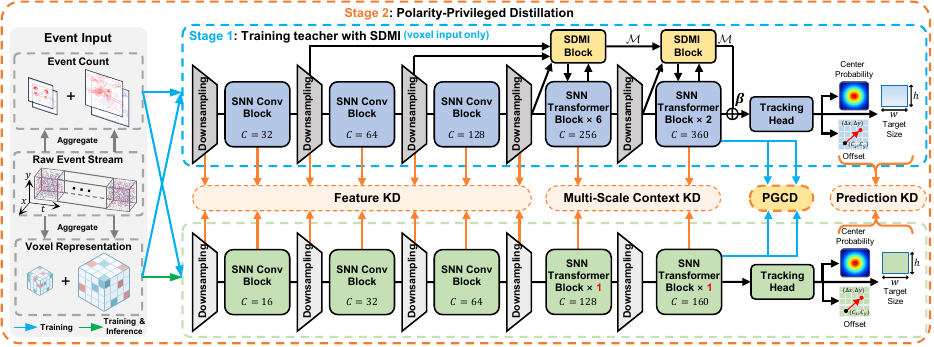}
  \caption{The two training stages. Stage 1 trains the teacher M-SDTrack,
which integrates SDMI, on the voxel representation alone. Stage 2 distils
it into the student EM-SDTrack through PPD, where both networks still take
the voxel representation as input and the training-only event counts serve
solely as the probe along which they are compared. At inference, the
student runs on the voxel representation alone.}
  \label{fig:framework}
\end{figure*}

\subsection{Distillation and privileged learning}
The PPD we propose builds on knowledge distillation, which transfers the
capability of a large model to a small one by matching
output distributions~\cite{hinton2015distilling}, intermediate
representations~\cite{zagoruyko2016paying}, inter-sample
relations~\cite{park2019relational} and attention
maps~\cite{jiao2020tinybert}. MixFormerV2 adds progressive layer deletion
to feature and logit distillation~\cite{cui2023mixformerv2}.
CompressTracker swaps student stages for teacher stages in
training~\cite{hong2025general}. Both give teacher and student the
same input. Privileged learning instead exploits training-only
information~\cite{vapnik2009new}. Hoffman et al.
hallucinate privileged-modality features from
RGB~\cite{hoffman2016learning}, Lambert et al. set the dropout variance
from it~\cite{lambert2018deep} and generalised distillation distils a
teacher trained on it~\cite{lopez2015unifying}. The PPD we propose takes
the polarity counts cancelled by the voxel
representation~\cite{yang2026fully} as privileged information. Whereas
the above let one network use it as evidence, PPD feeds it to both
networks only as the probe along which they are compared, and PGCD
transfers the curvature of the teacher's template--search relation along
this probe.

\section{Method}

\subsection{Overview}

EM-SDTrack is obtained in two training stages (Fig.~\ref{fig:framework}).
Stage~1 trains M-SDTrack, which integrates SDMI, with the standard tracking
loss, and the resulting model requires no floating-point multiplication at
inference. Stage~2 takes the trained M-SDTrack as the teacher and distils
it through PPD into EM-SDTrack, which has fewer channels and layers.

\subsection{Spiking neuron and spike-driven inference}
\label{sec:prelim}

Spiking neurons simulate the integrate-leak-fire process of biological
neurons. We describe this process as
\begin{align}
\mathbf{U}[t] &= \tau \mathbf{H}[t - 1] + \mathbf{X}[t], \label{eq4} \\
\mathbf{S}[t] &= \mathrm{Clip}\left(\mathrm{round}\left(\mathbf{U}[t]\right), 0, D\right), \label{eq5} \\
\mathbf{H}[t] &= \mathbf{U}[t] - \mathbf{S}[t]. \label{eq6}
\end{align}
This formulation is the I-LIF neuron, which is widely adopted in SNNs~\cite{yao2025scaling,luo2025integer}.
$\mathbf{U}[t]$ is the membrane potential after the input $\mathbf{X}[t]$ at
time $t$, $\mathbf{S}[t]$ is the spike value given by Eq.~\eqref{eq5}, and
$\mathbf{H}[t]$ is the membrane potential after the spike. $\tau$ is the
decay factor,
$\mathrm{Clip}(x, \text{min}, \text{max})$ constrains $x$ to
$[\text{min}, \text{max}]$, and $\mathrm{round}(\cdot)$ performs rounding.

At inference, the spike-ahead principle converts the integer output of
this neuron into binary spikes by incorporating $D$ into the iterative
timestep $T$, which gives $T \times D$ timesteps. Multiply-accumulate (MAC) operations are thereby
replaced by accumulate (AC) operations over the spatial locations whose
spike value is 1, which is what makes spike-driven inference low-power.

\subsection{Spike-driven multi-scale interaction}
\label{sec:sdmi}

Following the voxel representation~\cite{yang2026fully}, an event stream
$\{e_i=(x_i,y_i,t_i,p_i)\}$ with polarity $p_i\in\{+1,-1\}$ is partitioned
into three temporal bins of equal duration. Within the $n$-th bin
$\mathcal{E}_n$, events of opposite polarity at the same pixel cancel, and
the magnitude of the residual,
$I_n(x,y)=\big|\sum_{e_i\in\mathcal{E}_n,\,(x_i,y_i)=(x,y)}p_i\big|$,
is the pixel intensity. The three bins form a three-channel voxel frame,
from which the template $\mathbf{z}$ and the search sequence $\mathbf{X}$
are cropped, and the tracker localises the target of $\mathbf{z}$ in every
frame of $\mathbf{X}$.

M-SDTrack, shown in the upper part of Fig.~\ref{fig:framework}, processes
the template $\mathbf{z}$ and a search frame $\mathbf{x}$ of $H\times W$
pixels jointly. Let $\mathbf{F}_1$, $\mathbf{F}_2$ and $\mathbf{F}_3$ denote
the features produced by the three convolutional stages. Each is split into
a template part and a search part,
$\mathbf{F}_k=[\mathbf{F}_k^{\mathbf{z}};\mathbf{F}_k^{\mathbf{x}}]$, and
the search parts have spatial sizes $\tfrac{H}{4}\times\tfrac{W}{4}$,
$\tfrac{H}{8}\times\tfrac{W}{8}$ and $\tfrac{H}{16}\times\tfrac{W}{16}$,
respectively. SDMI acts on the search part only and maintains a Multi-Scale
Latent Buffer $\mathcal{M}$ built from $\mathbf{F}_1^{\mathbf{x}}$,
$\mathbf{F}_2^{\mathbf{x}}$ and $\mathbf{F}_3^{\mathbf{x}}$
(Sec.~\ref{sec:buffer}). Before each SNN Transformer stage, the search
tokens are enriched with context read from $\mathcal{M}$. After the stage,
the search part of its output is written back to $\mathcal{M}$. The template
tokens bypass SDMI, so the template remains an unmodified reference while
multi-scale context is injected where the target scale varies. With $\mathcal{T}_s=\mathcal{B}_s^{(L_s)}\circ\cdots\circ\mathcal{B}_s^{(1)}$
denoting the $s$-th stage, the composition of its $L_s$ SNN Transformer
blocks, the output of the first stage is
\begin{equation}
\mathbf{F}_4=\mathcal{T}_1\!\left(\left[\mathbf{F}_3^{\mathbf{z}};\,
\mathrm{SDMI}_1\!\left(\mathbf{F}_3^{\mathbf{x}};\mathcal{M}\right)\right]\right),
\label{eq:f4}
\end{equation}
and the second stage is applied analogously. Finally, $\mathcal{M}$ is
aggregated to the output resolution and added to the backbone output
(Sec.~\ref{sec:buffer}).

Every linear and convolutional layer of SDMI acts on spikes, as those of
the backbone do (Sec.~\ref{sec:prelim}), and its sampling read, DSS, costs
only shifts and additions (Sec.~\ref{sec:dss}). The network therefore
remains spike-driven at inference with SDMI integrated.

\subsubsection{SDMI block}
\label{sec:buffer}

The search parts of the three convolutional features are flattened and
brought to a common width $C$ by a spiking linear layer. Concatenating
them along the token axis then yields the Multi-Scale Latent Buffer
\begin{equation}
\mathcal{M}=\big[\mathbf{M}_1;\mathbf{M}_2;\mathbf{M}_3\big],\qquad
\mathbf{M}_k=\phi_k\!\big(\mathbf{F}_k^{\mathbf{x}}\big)\in\mathbb{R}^{N_k\times C},
\label{eq:buffer}
\end{equation}
where $\phi_k$ denotes this projection and $N_k$ is the number of tokens of
$\mathbf{F}_k^{\mathbf{x}}$. We refer to $\mathbf{M}_k$ as the $k$-th scale
of the buffer, the third of which has the resolution
$\tfrac{H}{16}\times\tfrac{W}{16}$ of the Transformer tokens. $\mathcal{M}$
is built once and persists across both Transformer stages. Between them, a
spiking linear layer maps it to the width of the second stage.

\begin{figure}[t]
  \centering
  \includegraphics[width=0.45\textwidth]{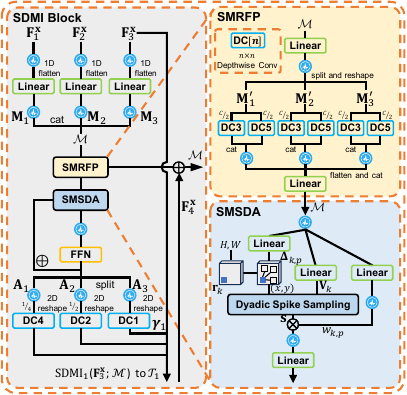}
  \caption{The SDMI block, drawn for the first Transformer stage. The upper
part builds the buffer $\mathcal{M}$ once. The second block reuses the
buffer returned by the first block, after a spiking linear layer maps it
to the width of the second stage.}
  \label{fig:sdmi}
\end{figure}

An SDMI block precedes each stage and receives the current search tokens
$\mathbf{Q}$, with $\mathbf{Q}=\mathbf{F}_3^{\mathbf{x}}$ for the first
stage. Its two modules, a Spiking Multi-Receptive
Field Feature Pyramid (SMRFP) and a Spiking Multi-Scale Deformable
Aggregation (SMSDA), are expanded on the right of Fig.~\ref{fig:sdmi}. SMRFP refines
the buffer at every scale with a bottleneck of two linear layers around
depthwise $3\times3$ and $5\times5$ convolutions on complementary
channel halves. SMSDA then lets every buffer token gather $P$ samples from
each scale at offsets predicted from the token itself, read with DSS
(Sec.~\ref{sec:dss}) and combined with weights emitted by a spiking neuron
instead of a softmax, and a spiking feed-forward layer follows. With $\mathbf{A}_k$ the part of the result at scale $k$, the enriched
search tokens returned by the block are
\begin{equation}
\mathrm{SDMI}_s(\mathbf{Q};\mathcal{M})
=\mathbf{Q}+\boldsymbol{\gamma}_s\odot\sum_{k=1}^{3}\mathcal{D}_k(\mathbf{A}_k),
\label{eq:sdmi_read}
\end{equation}
where $\mathcal{D}_k$ is a spiking depthwise convolution that brings scale
$k$ to the resolution of the Transformer tokens. $\boldsymbol{\gamma}_s$ is
a channel-wise gate initialised near zero, so that each stage starts from
its plain input and learns how much multi-scale context to admit. Being
channel-wise, $\boldsymbol{\gamma}_s$ rescales the outputs of
$\mathcal{D}_k$ and is folded into their weights at inference. The same
holds for $\boldsymbol{\beta}$ and $\mathcal{D}'_k$ below.

After the stage, the search part of its output is added to the third scale
of $\mathcal{M}$. The next block and the final aggregation thus see context
that has passed through the Transformer. Denoting the output of the second
stage by $\mathbf{F}_5$, the buffer is brought to the output resolution in
the same way and added through a learnable channel-wise gate
$\boldsymbol{\beta}$,
\begin{equation}
\mathbf{F}_{\mathrm{out}}^{\mathbf{x}}
=\mathbf{F}_5^{\mathbf{x}}+\boldsymbol{\beta}\odot\sum_{k=1}^{3}\mathcal{D}'_k(\mathbf{M}_k).
\label{eq:sdmi_out}
\end{equation}
This explicit path supervises the multi-scale representation by the
tracking loss directly and preserves the context written back to
$\mathcal{M}$ after the final stage.

\subsubsection{Dyadic spike sampling}
\label{sec:dss}

SMSDA chooses its sampling positions as deformable
attention~\cite{zhu2020deformable} does in ViT-Adapter~\cite{chen2022vision},
so that a token reads a few values from each scale at positions it predicts
itself instead of attending to every position. Let $N=N_1+N_2+N_3$ be the number of buffer
tokens and $P=4$ the number of samples per scale; attention heads are
omitted. Every token owns a reference point on each of the three scales,
$\mathbf{r}\in\mathbb{R}^{N\times3\times2}$, which is its own position
mapped to the $H_k\times W_k$ grid of scale $k$ ($H_kW_k=N_k$), and
predicts from its own features $P$ sampling offsets per scale,
$\boldsymbol{\Delta}\in\mathbb{R}^{N\times3\times P\times2}$, in pixels of
that grid. $\mathbf{r}_k$ and $\boldsymbol{\Delta}_{k,p}$ denote the entries
of one token. A spiking linear layer turns each scale $\mathbf{M}_k$ into a
value map $\mathbf{V}_k\in\mathbb{R}^{C\times H_k\times W_k}$, which the
token reads at the $P$ positions $\mathbf{r}_k+\boldsymbol{\Delta}_{k,p}$
of every scale, summing the $3P$ samples with weights $w_{k,p}$ emitted by
a spiking neuron (Fig.~\ref{fig:sdmi}). Offsets, weights and values thus
come from spiking linear layers, and the sum multiplies samples by spikes
only. The one operation left open is the read of $\mathbf{V}_k$ between
grid points, which DSS performs with shifts and additions only.

DSS reads $\mathbf{V}_k$ at $(x,y)=\mathbf{r}_k+\boldsymbol{\Delta}_{k,p}$
from the four grid values around $(x,y)$, zero outside the map, after
rounding the fractional part of the position to $b$ bits
(Algorithm~\ref{alg:dss}; Appendix~A.1 derives each line). Every
weight is then an integer over a power of two, so nothing beyond shifts and
additions is needed. In the algorithm,
$\mathbf{a}\ll t$ and $\mathbf{a}\gg t$ shift every entry of $\mathbf{a}$ by
$t$ bits, i.e.\ multiply or divide it by $2^{t}$, and
$q\triangleright\mathbf{d}=\sum_{t\in\mathrm{bits}(q)}(\mathbf{d}\ll t)=q\mathbf{d}$
multiplies a vector $\mathbf{d}$ by an integer $q$ with one shifted copy of
$\mathbf{d}$ per one in the binary representation of $q$; for $q=5=101_2$,
for instance, $q\triangleright\mathbf{d}=(\mathbf{d}\ll2)+\mathbf{d}$.

\begin{algorithm}[t]
\caption{Dyadic Spike Sampling}
\label{alg:dss}
\begin{algorithmic}[1]
\REQUIRE value map $\mathbf{V}$, position $(x,y)$, bits $b$
\ENSURE sample $\mathbf{s}\in\mathbb{R}^{C}$
\STATE $(x_0,y_0)\gets(\lfloor x\rfloor,\lfloor y\rfloor)$
\STATE $(q_x,q_y)\gets\mathrm{round}\big((x-x_0,\,y-y_0)\ll b\big)$
\STATE $\mathbf{V}_{ij}\gets\mathbf{V}[:,y_0+j,x_0+i]$ for $i,j\in\{0,1\}$
\STATE $\mathbf{h}_j\gets(\mathbf{V}_{0j}\ll b)
  +q_x\triangleright(\mathbf{V}_{1j}-\mathbf{V}_{0j})$ for $j\in\{0,1\}$
\STATE $\mathbf{s}\gets\big((\mathbf{h}_0\ll b)
  +q_y\triangleright(\mathbf{h}_1-\mathbf{h}_0)\big)\gg2b$
\RETURN $\mathbf{s}$
\end{algorithmic}
\end{algorithm}

\begin{proposition}[Dyadic weights]
\label{prop:dss}
For any $b\ge1$: (i) Algorithm~\ref{alg:dss} returns a weighted sum of
the four grid values,
$\mathbf{s}=2^{-2b}\sum_{i,j}\kappa_{ij}\mathbf{V}_{ij}$ with integer
weights $\kappa_{ij}\ge0$ summing to $2^{2b}$, so they are normalised
exactly, and uses at most $3b+3$ additions and $3b+4$ shifts per
channel and no multiplication. (ii) Let $\mathbf{s}_{\mathrm{cont}}$ be
the sample without the rounding in line~2, and let
$\Lambda_h=\max_j\|\mathbf{V}_{1j}-\mathbf{V}_{0j}\|_\infty$ and
$\Lambda_v=\max_i\|\mathbf{V}_{i1}-\mathbf{V}_{i0}\|_\infty$ be the largest
horizontal and vertical differences between adjacent grid values; then
\begin{equation}
\|\mathbf{s}-\mathbf{s}_{\mathrm{cont}}\|_\infty
\le 2^{-(b+1)}\,(\Lambda_h+\Lambda_v).
\label{eq:dss_bound}
\end{equation}
\end{proposition}
\noindent Appendix~A.1 proves part (i) and Appendix~A.2
part (ii). The bound halves with every additional bit and depends only on
how much the value map changes between neighbouring grid points. We use $b=8$, for which
it equals $(\Lambda_h+\Lambda_v)/512$. Training uses the dyadic sample in
the forward pass and the gradient of $\mathbf{s}_{\mathrm{cont}}$, i.e.\ of
standard bilinear interpolation, in the backward pass, since the rounding in
line~2 has no gradient. Under this straight-through estimator the offsets
are learned under the quantisation they meet at inference, so
Eq.~\eqref{eq:dss_bound} is a worst case rather than the accuracy actually
lost (Sec.~\ref{abla}).

\noindent\textbf{Spike-driven inference.}
Every scalar factor in SMSDA is thus a stored weight, a spike or, in the
DSS read, a power of two (Proposition~\ref{prop:dss}(i)), and the gates
are folded into the depthwise convolutions (Sec.~\ref{sec:buffer}). M-SDTrack therefore
runs at inference on accumulate, shift and add operations and contains no
floating-point multiplication.

\subsection{Polarity-privileged distillation}
\label{sec:ppd}

Stage~2 distils the trained M-SDTrack into the more efficient EM-SDTrack
(lower part of Fig.~\ref{fig:framework}). The student keeps the
teacher's topology at about half the channel width, retains two Transformer blocks after
progressive shrinking~\cite{cui2023mixformerv2}, shrinks the head
accordingly and contains no SDMI module. The voxel representation~\cite{yang2026fully}, which underlies the
best-performing event single-object trackers (Appendix~A.3),
however, cancels positive and negative events at the same pixel, and a
student with less context than the teacher cannot resolve the resulting
ambiguity. PPD therefore takes the polarity statistics discarded by the cancellation as
privileged information (Sec.~\ref{sec:privileged}), builds the constraint
PGCD on them (Sec.~\ref{sec:pgcd}) and combines it with the standard
distillation losses on features and predictions (Sec.~\ref{sec:objective}).

\subsubsection{Polarity as privileged information}
\label{sec:privileged}

Let $c^{+}$ and $c^{-}$ be the numbers of positive and negative events a
pixel receives within a bin. The voxel frame of Sec.~\ref{sec:sdmi} keeps
only $I_n=|c^{+}-c^{-}|$. One positive event and ten positive with nine
negative events therefore give the same intensity, and the cancelled part,
$2\min(c^{+},c^{-})$, is lost. Cancellation is deliberate and useful at
inference, since it suppresses noise and event-rate bias. Its collisions,
however, fall exactly where a tracker needs evidence. They arise for a small
or fast target whose leading and trailing edges land in the same cell, for
a target and a distractor with equal net count but different activity, and
for a target whose net polarity flips between template and search.

\begin{figure}[t]
  \centering
  \includegraphics[width=0.45\textwidth]{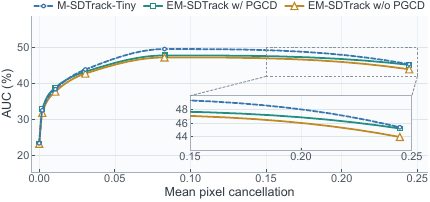}
  \caption{AUC on VisEvent versus the mean pixel cancellation, the first
factor of Eq.~\eqref{eq:gate} averaged over the pixels of the ground-truth
box. The frames of all test sequences form one bin for zero values and five
quantile bins for positive values.}
  \label{fig:gap}
\end{figure}

Neither network uses $c^{+}$ or $c^{-}$ for its decision. Once cancellation
has removed the pixel-level evidence, such a region can be resolved only
from its surroundings, that is, from the structure of the moving edge at
other scales. The teacher gathers this structure through SDMI, whereas the
smaller student, which has no SDMI module, cannot. Without PGCD, the
teacher--student gap therefore concentrates where cancellation occurs
(Fig.~\ref{fig:gap}). PPD thus treats the pre-cancellation counts as
privileged information~\cite{vapnik2009new}, available during training and
absent at test time. Both networks take them as input during training, but neither uses them
as evidence for its decision, unlike the teacher of generalised
distillation~\cite{lopez2015unifying}. They only define the probe along
which the two networks are compared. A light adapter turns the counts of
template and search into guidance maps of the form
\begin{equation}
g=\frac{2\min(c^{+},c^{-})}{c^{+}+c^{-}}\;\bar{a}\;(1+\eta\,\psi),
\qquad \psi\in(-1,1),
\label{eq:gate}
\end{equation}
where $\bar{a}$ is the activity $\log(1+c^{+}+c^{-})$ normalised by its
maximum over the frame. $g$ is thus large only where both polarities are
strongly active and zero where no event arrives. $\psi$ is the output of a
four-layer convolutional adapter with a zero-initialised head, shared by
template and search, which refines the physical prior by at most a factor
$1\pm\eta$ ($\eta=0.3$). The maps $\mathbf{g}^{\mathbf{z}}$ and
$\mathbf{g}^{\mathbf{x}}$ mark the polarity-ambiguous regions, and
Sec.~\ref{sec:pgcd} uses them to decide where teacher and student are
compared and in which direction they are probed.

\subsubsection{Polarity-guided curvature distillation}
\label{sec:pgcd}

Standard distillation matches outputs or features of teacher and student
at the given input. In a polarity-ambiguous region, however, the input
carries little of the evidence that determines the template--search
correspondence, so matching values there transfers little. What the teacher
knows shows instead in how its correspondence changes when the evidence
there is strengthened or weakened, and PGCD transfers this behaviour by
probing teacher and student along the same polarity-guided direction.

The probe acts on $\mathbf{F}_3$, the joint template--search feature that
enters the Transformer stages, in which template and search tokens interact
(Sec.~\ref{sec:sdmi}).
The guidance maps of Sec.~\ref{sec:privileged} are resampled to the
resolution of $\mathbf{F}_3$ and concatenated like its template and search
parts into $\mathbf{g}$. The features of every token are then scaled up and
down in proportion to its guidance value,
\begin{equation}
\mathbf{F}_3^{\pm}=\mathbf{F}_3\odot(1\pm\epsilon\,\mathbf{g}),\qquad
\epsilon=0.15,
\label{eq:probe}
\end{equation}
and the base, plus and minus copies pass through the rest of the network
with shared weights, SDMI included for the teacher. We summarise the tokens
$\mathbf{F}_5$ of each copy (Sec.~\ref{sec:buffer}) by the template--search
relation
\begin{equation}
\mathbf{R}=\mathrm{softmax}\big(\hat{\mathbf{F}}_5^{\mathbf{x}}\,
\hat{\mathbf{F}}_5^{\mathbf{z}\top}/\vartheta\big),\qquad\vartheta=0.1,
\label{eq:relation}
\end{equation}
where the hat normalises every token to unit length and the softmax is
taken over each row. The entry $R_{ij}$ is thus the weight with which
search token $i$ is assigned to template token $j$. Being a
correspondence rather than a feature, the relation is comparable between
networks of different widths without any adapter.

The relation's central second difference along the probe,
\begin{equation}
\mathbf{K}=\mathbf{R}^{+}+\mathbf{R}^{-}-2\mathbf{R}^{0},
\label{eq:curvature}
\end{equation}
cancels the first-order change by symmetry and retains how the
correspondence bends when the ambiguous evidence is pushed either way, which
we call the curvature of the relation. PGCD matches the curvatures of
student and teacher,
\begin{equation}
\mathcal{L}_{\mathrm{PGCD}}=\frac{\sum_{i,j}\mu_{ij}\,
\ell\big(K^{\mathsf{S}}_{ij}-K^{\mathsf{T}}_{ij}\big)}
{\sum_{i,j}\mu_{ij}},
\label{eq:pgcd}
\end{equation}
where $\ell$ is the smooth-$L_1$ loss with threshold $0.4$ and
$\mathbf{K}^{\mathsf{T}}$ is detached. The weight of a pair,
$\mu_{ij}=1+g^{\mathbf{x}}_i+g^{\mathbf{z}}_j$ with $g^{\mathbf{x}}_i$ and
$g^{\mathbf{z}}_j$ the guidance values at search token $i$ and template
token $j$, grows when either token lies in a polarity-ambiguous region. The
division by the sum of the weights keeps the scale of the loss independent of how much of a frame is
ambiguous. The adapter of Sec.~\ref{sec:privileged} is trained jointly with
the student through this loss.

\subsection{Training objective}
\label{sec:objective}
The teacher is trained in Stage~1 with the standard tracking loss
$\mathcal{L}_{\mathrm{trk}}$, in which the focal loss on the centre map
and the $L_1$ and GIoU losses on the box are weighted by $1$, $5$ and $2$.
In Stage~2 the teacher is frozen and the student minimises
% Continuation lines are indented by a quad (as in the appendix inequality
% eq:dss_triangle) and \! precedes each sum, so that the wider PKD term
% leaves room for the equation number on the middle line.
\begin{equation}
\begin{aligned}
&\mathcal{L}=\lambda_{\mathrm{gt}}\,\mathcal{L}_{\mathrm{trk}}
+\lambda_{\mathrm{kd}}\big(\mathcal{L}_{\mathrm{box}}
+\|\mathbf{Y}^{\mathsf{S}}-\mathbf{Y}^{\mathsf{T}}\|_2^2\big)\\
&\quad+\lambda_{\mathrm{c}}\!\sum_{l\in\mathrm{conv}}
\mathcal{L}_{\mathrm{PKD}}\big(\tilde{\mathbf{f}}^{\mathsf{S}}_l,
\mathbf{f}^{\mathsf{T}}_l\big)
+\lambda_{\mathrm{a}}\!\sum_{l\in\mathrm{attn}}
\|\tilde{\mathbf{f}}^{\mathsf{S}}_l-\mathbf{f}^{\mathsf{T}}_l\|_2^2\\
&\quad+\lambda_{\mathrm{o}}\,\|\tilde{\mathbf{F}}^{\mathsf{S}}_{\mathrm{out}}
-\mathbf{F}^{\mathsf{T}}_{\mathrm{out}}\|_2^2
+\lambda_{\mathrm{p}}\,\mathcal{L}_{\mathrm{PGCD}},
\end{aligned}
\label{eq:objective}
\end{equation}
where $\mathcal{L}_{\mathrm{box}}$ applies the $L_1$ and GIoU losses
between the boxes of student and teacher, $\mathbf{Y}^{\mathsf{S}}$ and
$\mathbf{Y}^{\mathsf{T}}$ are their centre maps,
$\mathbf{f}_l$ is the output of layer $l$, a tilde marks a
student feature mapped to the width of the teacher by a linear adapter,
and $\mathbf{F}_{\mathrm{out}}$ is the backbone output.
$\mathcal{L}_{\mathrm{PKD}}$ is the PKD loss~\cite{cao2022pkd}, which
maximises the Pearson correlation between its arguments. The two sums run
over the intermediate layers of the convolutional stages (feature KD) and
over the Transformer blocks, whose outputs in the teacher carry the
multi-scale context injected by SDMI (multi-scale context KD,
Fig.~\ref{fig:framework}). The weights are $\lambda_{\mathrm{gt}}=0.7$,
$\lambda_{\mathrm{kd}}=1$, $\lambda_{\mathrm{c}}=0.08$,
$\lambda_{\mathrm{a}}=0.13$, $\lambda_{\mathrm{o}}=0.09$ and
$\lambda_{\mathrm{p}}=0.6$ in all experiments.

\begin{table*}[!htbp]
\centering
\setlength{\tabcolsep}{5pt}
\small
\caption{Comparison with ANN-based and SNN-based trackers on FE108, FELT
and VisEvent. Energy is estimated theoretically. The best
three results are shown in \textcolor{red}{\textbf{red}},
\textcolor{blue}{\textbf{blue}} and \textcolor{papergreen}{\textbf{green}}.}
\begin{tabular}{@{}ccccccccccc@{}}
\toprule
\multirow{2}{*}{Methods}
  & \multirow{2}{*}{\begin{tabular}[c]{@{}c@{}}Param.\\ (M)\end{tabular}}
  & \multirow{2}{*}{\begin{tabular}[c]{@{}c@{}}Spiking \\ Neuron\end{tabular}}
  & \multirow{2}{*}{\begin{tabular}[c]{@{}c@{}}Timesteps\end{tabular}}
  & \multirow{2}{*}{\begin{tabular}[c]{@{}c@{}}Energy\\ (mJ)\end{tabular}}
  & \multicolumn{2}{c}{FE108}
  & \multicolumn{2}{c}{FELT}
  & \multicolumn{2}{c}{VisEvent}
\\
\cmidrule(l){6-11}
  & & & &
    & AUC(\%) & PR(\%)
    & AUC(\%) & PR(\%)
    & AUC(\%) & PR(\%)
\\ \midrule
STARK~\cite{yan2021learning}           & 28.23  & –      & 1 & 58.88  & 57.4 & 89.2 & 39.3 & 50.8 & 34.1 & 46.8 \\
SimTrack~\cite{chen2022backbone}      & 88.64  & –      & 1 & 93.84  & 56.7 & 88.3 & 36.8  & 47.0  & 34.6 & 47.6 \\
OSTrack\textsubscript{256}~\cite{ye2022joint}         & 92.52  & –      & 1 & 98.90  & 54.6 & 87.1 & 35.9  & 45.5  & 32.7 & 46.4 \\
ARTrack\textsubscript{256}~\cite{wei2023autoregressive}
                                        & 202.56 & –      & 1 & 174.80 & 56.6 & 88.5 & 39.5 & 49.4 & 33.0 & 43.8 \\
SeqTrack-B\textsubscript{256}~\cite{chen2023seqtrack}
                                        & 90.60  & –      & 1 & 302.68 & 53.5 & 85.5 & 33.0  & 42.0  & 28.6 & 43.3 \\
HiT-B~\cite{kang2023exploring}           & 42.22  & –      & 1 & 19.78  & 55.9 & 88.5 & 38.5  & 48.9  & 34.6 & 47.6 \\
GRM~\cite{gao2023generalized}           & 99.83  & –      & 1 & 142.14 & 56.8 & 89.3 & 37.2  & 47.4  & 33.4 & 47.7 \\
HIPTrack~\cite{cai2024hiptrack}         & 120.41 & –      & 1 & 307.74 & 50.8 & 81.0 & 38.2  & 48.9  & 32.1 & 45.2 \\
ODTrack~\cite{zheng2024odtrack}         & 92.83  & –      & 1 & 335.80 & 43.2 & 69.7 & 29.7  & 35.9  & 24.7 & 34.7 \\
% SiamRPN~\cite{li2018high}               & –      & –      & 1 & –      & –    & –    & –     & –     & 24.7 & 38.4 \\
% ATOM~\cite{danelljan2019atom}           & –      & –      & 1 & –      & –    & –    & 22.3  & 28.4  & 28.6 & 47.4 \\
% DiMP~\cite{bhat2019learning}           & –      & –      & 1 & –      & –    & –    & 37.8  & 48.5  & 31.5 & 44.2 \\
% PrDiMP~\cite{danelljan2020probabilistic}
%                                           & –      & –      & 1 & –      & –    & –    & 34.9  & 44.5  & 32.2 & 46.9 \\
% MixFormer~\cite{cui2022mixformer}      & 37.55  & –      & 1 & –      & –    & –    & 38.9  & 50.4  & –    & –    \\
\midrule
STNet~\cite{zhang2022spiking}         & \textcolor{papergreen}{\textbf{20.55}}  & LIF    & 3 & 103.53 & –    & –    & –     & –     & 35.0 & 50.3 \\
SNNTrack~\cite{zhang2025spiking}      & 31.40  & BA-LIF & 5 & 8.25   & –    & –    & –     & –     & 35.4 & 50.4 \\
SDTrack-Tiny~\cite{shan2026sdtrack}
    & \textcolor{blue}{\textbf{19.61}}  & I-LIF  & 4 & \textcolor{papergreen}{\textbf{8.16}}
    & 59.0 &91.3 & 39.3 & 51.2 & 35.6 & 49.2 \\
SDTrack-Base~\cite{shan2026sdtrack}
    & 107.26 & I-LIF  & 4 & 30.52
    & 59.9 & 91.5 & 40.0 & 51.4 &37.4 & 51.5 \\
SpikeET~\cite{yang2026fully}      & 22.36  & I-LIF  & 4 &  11.46   & \textcolor{papergreen}{\textbf{63.9}} &93.7 & – & – &39.9  &54.8 \\
% BSA~\cite{wang2026bipolar}      & 19.61  & I-LIF  & 4 &        & 59.2 & 91.4 & 40.9 & 51.8 & 36.8 & 52.3 \\
MLPixer~\cite{zhang2026unveiling}      & 22.99  & LIF    & 4 &  –      & 57.9 & 90.1 & – & – & 34.5 & 48.9 \\
AdaS~\cite{zhouadas}      & \textcolor{blue}{\textbf{19.61}}  & I-LIF  & 4 &    –    & 60.2 & 92.5 & – & – & 36.3 & 50.5  \\
TP-Spikformer~\cite{wei2026tp}      & –      & I-LIF  & 4 &   \textcolor{blue}{\textbf{6.65}}  & 59.0 & 91.2 & 39.1 & 50.4 & 35.3 & 49.7  \\
% IPL V2-Tiny~\cite{}      & 22.28  & I-LIF  & 4 &        & 63.4    & 93.6    & 40.7     & 51.7     & 40.4     & 55.2  \\
% IPL V2-Base~\cite{}      & 114.40 & I-LIF  & 4 &        & 64.4    & 94.7    & 41.1     & 52.2     & 40.7     & 55.4  \\
\midrule
\textbf{M-SDTrack-Tiny}      &   23.39     & I-LIF  & 4 &    12.82    & \textcolor{blue}{\textbf{64.0}}    & \textcolor{blue}{\textbf{95.0}}    & \textcolor{blue}{\textbf{41.2}}     & \textcolor{blue}{\textbf{52.0}}     & \textcolor{blue}{\textbf{41.5}}     & \textcolor{blue}{\textbf{56.1}}  \\
\textbf{M-SDTrack-Base}      &  118.67      & I-LIF  & 4 &   40.68     & \textcolor{red}{\textbf{65.1}}    & \textcolor{red}{\textbf{95.8}}    & \textcolor{red}{\textbf{41.7}}     & \textcolor{red}{\textbf{52.9}}     & \textcolor{red}{\textbf{42.0}}     & \textcolor{red}{\textbf{57.1}}   \\
\cmidrule(l){2-11}
\textbf{EM-SDTrack}      &   \textcolor{red}{\textbf{4.24}}     & I-LIF  & 4 &   \textcolor{red}{\textbf{3.42}}     & 63.4   &  \textcolor{papergreen}{\textbf{94.0}}   &  \textcolor{papergreen}{\textbf{40.5}}   & \textcolor{papergreen}{\textbf{51.5}} & \textcolor{papergreen}{\textbf{41.0}}    & \textcolor{papergreen}{\textbf{55.5}}  \\

\bottomrule
\end{tabular}

\label{main_table}
\end{table*}

\section{Experiments}
\subsection{Implementation details}
\label{sec:impl}
M-SDTrack inherits the backbone of SDTrack, with a spiking depthwise
separable convolution inserted before each SNN Transformer block. This
convolution, adopted from SpikeFET, strengthens the local inductive bias
of the backbone. Empirically, it also improves the multi-scale
representation substantially (Appendix~A.4). EM-SDTrack,
shown in the lower part of Fig.~\ref{fig:framework}, retains this
backbone design at the reduced width and depth specified in
Sec.~\ref{sec:ppd}.

Training proceeds in two stages after pre-training. We first pre-train
the backbone of M-SDTrack, without SDMI, on ImageNet-1K with the recipe
of SDTrack. In Stage~1, M-SDTrack is then fine-tuned on each benchmark
through the pair-matching
task~\cite{yan2021learning,chen2022backbone,ye2022joint,kang2023exploring},
with 60,000 template--search pairs sampled per epoch. Fine-tuning runs
for 55 epochs on FE108 and VisEvent and for 300 epochs on FELT, with a
batch size of 40. The optimiser is AdaNSD with an
initial learning rate of $6\times10^{-4}$ and cosine annealing~\cite{zhouadas}. In
Stage~2, the trained M-SDTrack-Tiny serves as the teacher, and
EM-SDTrack is trained for 120 epochs with the same sampling. This stage
uses AdamW with an initial learning rate of $5\times10^{-4}$ and cosine
decay. Between epochs 10 and 20, the blocks of the two SNN Transformer
stages are progressively removed by cosine-decayed
dropping~\cite{cui2023mixformerv2} until one block per stage remains.
Both stages use the standard data augmentation
of~\cite{yang2026fully,wang2023visevent}.

Both stages are implemented in PyTorch~2.12 with CUDA~13.0 and run on
four NVIDIA H100 GPUs. For a fair comparison, the evaluation metrics and
the estimation of theoretical energy consumption follow SDTrack.

% Combined ablation table (2026-11-06): three sub-tables side by side,
% (a) SDMI components, (b) PPD objective, (c) retained Transformer blocks.
% The source sits in Sec. 4.2 (page 7) so that this double-column float
% lands at the top of the ablation page (page 8).
\begin{table*}[t]
\centering
\footnotesize
\caption{Ablation studies on FE108. (a)~Components of SDMI. (b)~Components
of PPD. Row~2 uses the tracking loss alone, rows 3--7 set one weight of
Eq.~\eqref{eq:objective} to zero in turn, and row~8 removes the surplus
Transformer blocks before training instead of dropping them progressively.
(c)~Index $l_s$ of the block kept in Transformer stage $s$, out of six
and two. Row~1 of each sub-table is the default.}
\setlength{\tabcolsep}{3pt}
\begin{subtable}[t]{0.36\textwidth}
\centering

\begin{tabular}{@{}cccc@{}}
\toprule
\# & Method & AUC(\%) & PR(\%) \\ \midrule
1 & \textbf{M-SDTrack-Tiny (baseline)}                & \textbf{64.0} & \textbf{95.0} \\
2 & Remove SDMI                              & 63.4 & 93.6 \\
3 & Remove SDMI\textsubscript{1}             & 63.8 & 94.5 \\
4 & Remove SDMI\textsubscript{2}             & 63.7 & 94.2 \\
5 & Remove SMRFP                             & 63.7 & 94.4 \\
6 & Remove SMSDA                             & 63.6 & 93.9 \\
7 & Remove Final Aggregation                 & 63.8 & 94.8 \\
8 & DSS $\rightarrow$ Bilinear Interpolation & 64.0 & 95.0 \\ \bottomrule
\end{tabular}
\caption{SDMI}
\label{tab:ablation_sdmi}
\end{subtable}
\hfill
\begin{subtable}[t]{0.42\textwidth}
\centering
\begin{tabular}{@{}cccc@{}}
\toprule
\# & Method & AUC(\%) & PR(\%) \\ \midrule
1 & \textbf{EM-SDTrack (distilled by PPD)}                      & \textbf{63.4} & \textbf{94.0} \\
2 & Train EM-SDTrack Directly                          & 51.9 & 84.5 \\
3 & Remove Prediction KD                               & 62.6 & 92.8 \\
4 & Remove Feature KD                                  & 62.3 & 92.1 \\
5 & Remove Multi-Scale Context KD                      & 61.8 & 92.6 \\
6 & Remove Backbone-Output KD                          & 62.9 & 93.7 \\
7 & Remove PGCD                                        & 62.5 & 92.8 \\
8 & Progressive Shrinking $\rightarrow$ Direct Removal & 63.0 & 93.4 \\ \bottomrule
\end{tabular}
\caption{PPD}
\label{tab:ablation_ppd}
\end{subtable}
\hfill
\begin{subtable}[t]{0.18\textwidth}
\centering
\begin{tabular}{@{}cccc@{}}
\toprule
$l_1$ & $l_2$ & AUC(\%) & PR(\%) \\ \midrule
\textbf{1} & \textbf{1} & \textbf{63.4} & \textbf{94.0} \\
2 & 1 & 63.2 & 93.6 \\
3 & 1 & 63.1 & 93.4 \\
4 & 1 & 62.8 & 92.9 \\
5 & 1 & 62.6 & 92.6 \\
6 & 1 & 62.2 & 92.2 \\
1 & 2 & 62.4 & 92.3 \\
6 & 2 & 62.0 & 91.9 \\ \bottomrule
\end{tabular}
\caption{Retained blocks}
\label{tab:ablation_blocks}
\end{subtable}

\label{tab:ablation}
\end{table*}

\begin{figure*}[t]
  \centering
  \begin{minipage}[t]{0.32\textwidth}
    \centering
    \includegraphics[width=\linewidth]{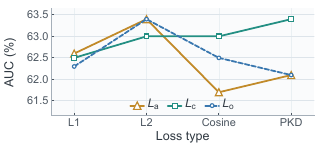}
    \par\vspace{2pt}
    {\small (a)}
  \end{minipage}\hfill
  \begin{minipage}[t]{0.32\textwidth}
    \centering
    \includegraphics[width=\linewidth]{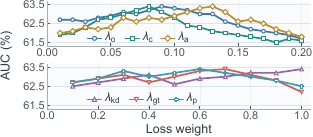}
    \par\vspace{2pt}
    {\small (b)}
  \end{minipage}\hfill
  \begin{minipage}[t]{0.32\textwidth}
    \centering
    \includegraphics[width=\linewidth]{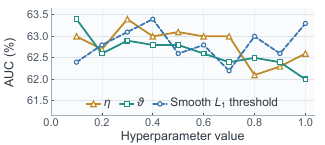}
    \par\vspace{2pt}
    {\small (c)}
  \end{minipage}
  \caption{Sensitivity of EM-SDTrack on FE108 to
  (a)~the loss functions of the terms $\mathcal{L}_{\mathrm{a}}$,
  $\mathcal{L}_{\mathrm{c}}$ and $\mathcal{L}_{\mathrm{o}}$ weighted by
  $\lambda_{\mathrm{a}}$, $\lambda_{\mathrm{c}}$ and $\lambda_{\mathrm{o}}$
  in Eq.~\eqref{eq:objective},
  (b)~the loss weights, and
  (c)~the internal hyperparameters of PPD:
  $\eta$, $\vartheta$ and the smooth-$L_1$ threshold.}
  \label{fig:sens}
\end{figure*}

\subsection{Results and comparisons}
We compare M-SDTrack and EM-SDTrack with representative ANN-based and
SNN-based trackers on FE108~\cite{zhang2021object}, FELT~\cite{wang2024long} and VisEvent~\cite{wang2023visevent} in
Tab.~\ref{main_table}. Accuracy is measured by the area under the success-plot curve
(AUC) and the precision rate (PR), and efficiency by the number of
parameters and the theoretical energy consumption. Differences in AUC
and PR are given in percentage points.

On FE108, M-SDTrack-Tiny exceeds SDTrack-Tiny by 5.0 in AUC and 3.7 in
PR, and M-SDTrack-Base exceeds SDTrack-Base by 5.2 and 4.3. These gains
cost 3.78\,M additional parameters at the Tiny scale. SDMI introduces
only 1.11\,M of them, and the remaining 2.67\,M belong to the
separable convolutions. SpikeET contains the same convolutions at a
comparable size but no cross-scale interaction.
M-SDTrack-Tiny still exceeds it by 1.3 in PR on FE108 and by 1.6 in AUC
and 1.3 in PR on VisEvent.

On the long-term benchmark FELT, the AUC gains over SDTrack are smaller
but consistent, 1.9 at the Tiny scale and 1.7 at the Base scale. On the
large-scale VisEvent benchmark, they grow to 5.9 and 4.6. M-SDTrack-Base
achieves state-of-the-art accuracy on all three benchmarks.

EM-SDTrack retains most of the accuracy of M-SDTrack-Tiny at a fraction
of its cost. It has about 18\% of the parameters and 27\% of the energy
consumption of this teacher, and keeps 99\% of the teacher's AUC and
PR (averaged over the three benchmarks and rounded to the nearest
integer). With 4.24\,M parameters, it exceeds every earlier SNN-based
tracker in Tab.~\ref{main_table} on FELT and VisEvent.

\subsection{Ablation study}
\label{abla}

\noindent\textbf{Effectiveness of SDMI.}
On FE108 with the protocol of Sec.~\ref{sec:impl}, every component of SDMI
contributes to the accuracy of M-SDTrack-Tiny (Tab.~\ref{tab:ablation}a).
Removing the final aggregation of Eq.~\eqref{eq:sdmi_out} in row~7 causes a
clear drop, since this path alone lets the tracking loss supervise the
buffer directly. DSS reaches accuracy comparable to standard bilinear
interpolation (row~8), so the bound of Eq.~\eqref{eq:dss_bound} is a worst
case rather than the accuracy actually lost. The initialisation of the
gates $\boldsymbol{\gamma}_s$ and $\boldsymbol{\beta}$ and the number of
bits $b$ of DSS are further ablated in Appendices~A.5
and~A.6.

\noindent\textbf{Effectiveness of PPD.}
Training EM-SDTrack with the tracking loss alone (row~2 of
Tab.~\ref{tab:ablation}b) falls short of the distilled model, which we
attribute to the limited capacity of its two Transformer blocks. Every
distillation term of Eq.~\eqref{eq:objective} contributes (rows 3--7), and
PGCD brings a clear gain (row~7), confined to targets with cancellation
(Fig.~\ref{fig:gap}).
Keeping the first block of each stage transfers its representation best
(Tab.~\ref{tab:ablation}c). The sensitivity to the distillation loss
functions, their weights and the internal hyperparameters of PPD is
analysed in Fig.~\ref{fig:sens}. The window of progressive shrinking is
further ablated in Appendix~A.7.

\begin{figure}[t]
  \centering
  \begin{minipage}[b]{0.48\linewidth}
    \centering
    \includegraphics[width=\linewidth]{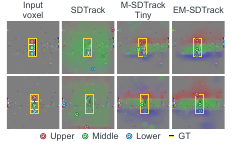}
    \par\vspace{2pt}
    {\small (a)}
  \end{minipage}\hfill
  \begin{minipage}[b]{0.48\linewidth}
    \centering
    \includegraphics[width=\linewidth]{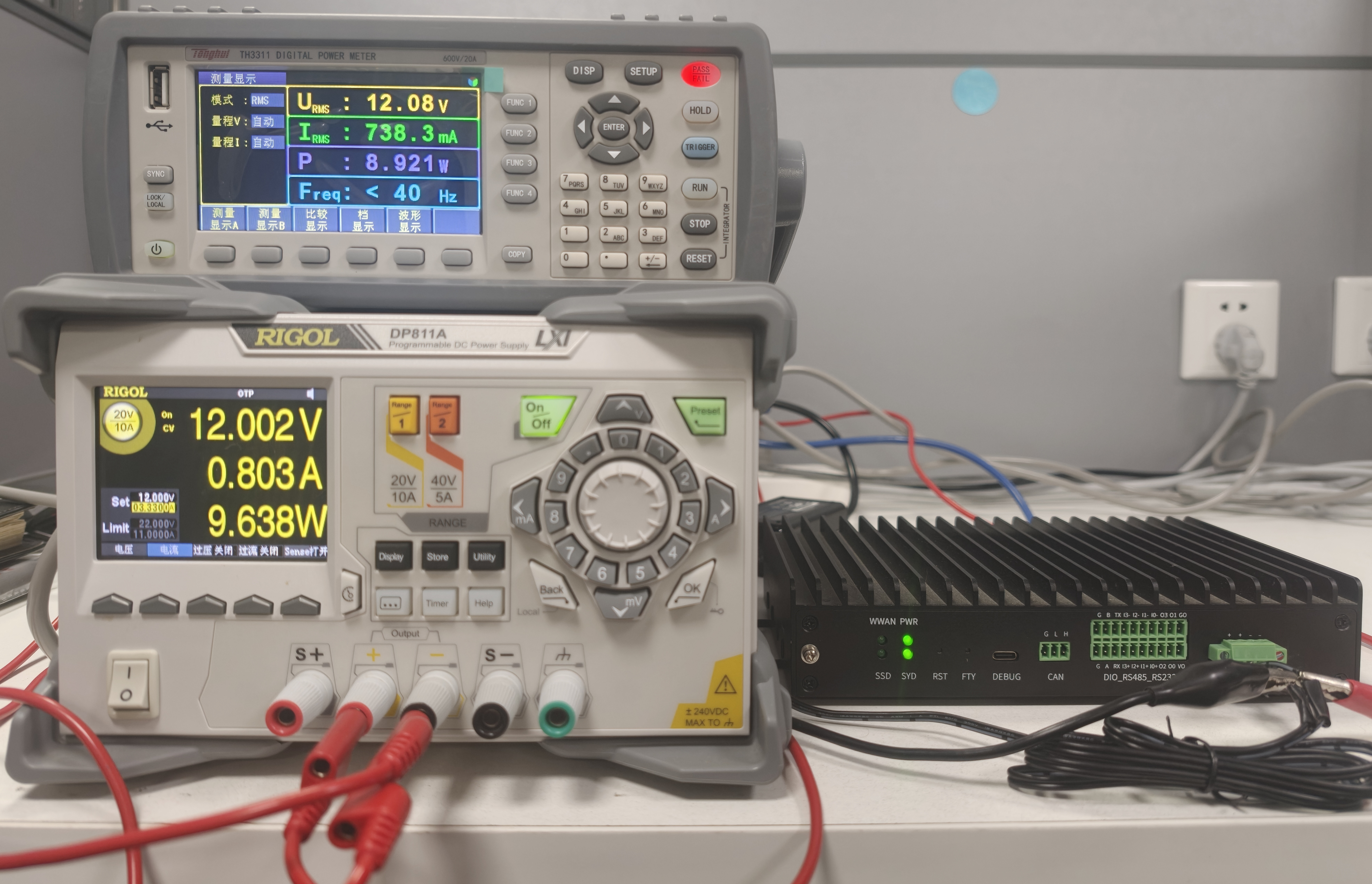}
    \par\vspace{2pt}
    {\small (b)}
  \end{minipage}
  \caption{(a)~Relation of Eq.~\eqref{eq:relation} averaged over the
  upper, middle and lower parts of the template target, on a colour scale
  shared by the three trackers. Circles mark the maxima. (b)~EM-SDTrack
  running on the Lynxi KA200 chip~\protect\footnotemark{} (right), with a programmable DC power supply (bottom) and a power meter (top).}
  \label{fig:deploy}
\end{figure}
\footnotetext{\url{https://www.lynxi.com/}}

\section{Discussion}

In Fig.~\ref{fig:deploy}a, samples from two sequences show that SDMI
enables M-SDTrack to locate the three parts of the target more accurately
than SDTrack. The maxima of SDTrack fall outside the ground-truth box for
all three parts. Those of M-SDTrack lie inside the box and follow the
vertical order of the parts. EM-SDTrack exhibits the same capability, which indicates that
PPD transfers it to the student. On the Lynxi KA200 neuromorphic chip
(Fig.~\ref{fig:deploy}b), EM-SDTrack runs in real time at 116~FPS, with
the same tracking accuracy as on a GPU. The total system power consumption
is 8.92~W, of which the chip accounts for 0.12~W.

\section{Conclusion}
This paper proposes SDMI and PPD for event-based spike-driven tracking under the
constraints of edge deployment. SDMI performs cross-scale interaction in a
single spiking forward pass. Its core, DSS, restricts sampling coordinates
to dyadic fractions, while the offsets are still learned with the gradient
of standard bilinear interpolation. M-SDTrack, which integrates SDMI, thus
remains fully spike-driven at inference and achieves state-of-the-art
accuracy on FE108, FELT and VisEvent. PPD distils M-SDTrack into
EM-SDTrack. It treats the event counts cancelled by the voxel
representation as privileged information and builds its core constraint,
PGCD, on them. EM-SDTrack retains 99\% of the accuracy of its teacher
M-SDTrack-Tiny with 18\% of its parameters and 27\% of its energy. On the Lynxi KA200 neuromorphic chip, it achieves lossless real-time inference at low power, laying the groundwork for neuromorphic tracking systems at the edge.

{
    \small
    \bibliographystyle{ieeenat_fullname}
    \bibliography{main}
}

\end{document}